\documentclass[10pt,twocolumn,letterpaper]{article}

\usepackage{appendix}
\usepackage{etoolbox}
\usepackage{times}
\usepackage{epsfig}
\usepackage{graphicx}
\usepackage{amsmath}
\usepackage{amssymb}
\usepackage[normalem]{ulem}
\usepackage[
    pagebackref=true
    breaklinks=true,
    letterpaper=true,
    colorlinks=true,
    bookmarks=false,
    urlcolor=blue,
    citecolor=black,
]{hyperref}

\BeforeBeginEnvironment{appendices}{\clearpage}

\title{\vspace{-2.0cm}Contrastive Learning in Distilled Models}
\author{
Valerie Lim \\ {\tt\small vlim31@gatech.edu}
\and
Kai Wen Ng \\ {\tt\small kng71@gatech.edu}
\and
Kenneth Lim \\ {\tt\small klim83@gatech.edu}
}

\date{\vspace{-5ex}}

\begin{document}

\maketitle

\begin{abstract}
    Natural Language Processing (NLP) models like BERT can provide state of the art word embeddings for downstream NLP tasks. However, these models: (1) yet to perform well on Semantic Textual Similarity (STS), and (2) may be too large to be deployed as lightweight edge applications. We seek to apply a suitable contrastive learning method based on the SimCSE paper, to a model architecture adapted from a knowledge distillation based model, DistilBERT, to address these two issues. Our final lightweight model DistilFACE achieves an average of 72.1 in Spearman’s correlation on STS tasks, a 34.2\% improvement over BERTbase with default parameters.
\end{abstract}

\section{Introduction}

\subsection{Problem Statement}

Pre-trained word embeddings are a vital part of natural language processing (NLP) systems, providing state of the art results for many NLP tasks  as the pre-trained embedding-based models frequently outperform alternatives such as the embedding-layer-based Word2Vec \cite{farahmand}. Accurate word embeddings have huge downstream benefits for tasks related to information retrieval and semantic similarity comparison \cite{2105.11741}, especially in a typical industrial setting where labelled data is scarce. As a result, in recent years, word embeddings from pre-trained models such as BERT (Bidirectional Encoder Representations from Transformers) have been gaining traction in many NLP works.

However, studies \cite{2104.08821} \cite{2105.11741} have shown that BERT does not produce good enough embeddings for semantic textual similarity tasks. As a result, recent breakthroughs \cite{2104.08821} \cite{1907.11692} have found that a contrastive objective is able to address the anisotropy issue of BERT. At a high level, contrastive learning is a technique where given two augmented samples of \(x\), \(x+\) (positive sample), \(x-\) (negative sample), we'd like for \(x+\) to be close to \(x\) and for \(x-\) to be far away from \(x\). This improves the quality of BERT sentence representation. 

Yet, a problem often associated with BERT and the contrastive loss models derived off it is their large size. There is an increasing need to build machine learning systems that can operate on the edge, instead of calling a cloud API, in order to build privacy respecting systems and prevent having users send sensitive information to servers \cite{10.3390/s22020450}. Thus, there is a growing impetus for lightweight and responsive models. However, most of the pre-trained models \cite{2104.08821} \cite{2105.11741} with contrastive loss are still relatively large and computationally heavy, making it hard to deploy on the edge. Deployment on the edge would also allow for reduced time and network latency, which could be critical in certain use cases such as when there is a high QOS (Quality of Service) requirement \cite{10.3390/s22020450}. A common solution for creating lighter models is via knowledge distillation, where a trained, complex teacher model trains  a lighter, shallower student model to generalise \cite{1910.01108}.
Hence, our goal was to address this joint problem of lack of lightweight models that can perform well on semantic textual similarity tasks, by creating an unsupervised pretrained model that combines knowledge distillation and contrastive learning to address their respective problems.  To do this, we first adopt the architecture and pre-trained checkpoints of a knowledge distillation based model, DistilBERT, start off with a lightweight model. We then adapt from authors who successfully implemented contrastive learning on sentence embeddings via unsupervised learning \cite{2104.08821}, by feeding samples, termed as \(x\), into the pre-trained encoder twice: by applying standard dropout twice to obtain two different sets of embeddings termed as \(x+\). Then we take other sentences in the same mini-batch as \(x-\) and the model predicts the positive sample among the negatives. We subsequently experiment with different pooling layers, automatic mixed precision, quantization and hyperparameter tuning to enhance performance.

\subsection{Who Cares? The Implications}

Our group believes there is significant value in experimenting with whether a traditional knowledge distillation method like DistilBERT, can be directly adapted to be combined with an existing contrastive learning method such as SimCSE, without having to create an entirely new model. Such an approach makes such a combination of techniques more accessible. In addition, there are definitely environmental and economical benefits to doing so, given how computational intensive large-scale pretraining of new models are. For instance, 16 TPU chips were required to train BERTbase, with each round of pretraining taking 4 days \cite{1810.04805}, costing about US\$6,912 each time \cite{cost.sota.ai}. Furthermore, success in our project will enable the creation of lightweight pre-trained embedding-based models that work well in semantic textual similarity tasks. This has many applications, such as searching and ranking of relevant documents based on similarity \cite{robust.sts}, and now it could be even further applied in edge scenarios. For example, such functionality can now be enabled at sensitive government, defence or health institutions, where computers are not always allowed to be connected to the cloud freely \cite{1910.01108}.

\section{How Is It Done Today and Limits}

\subsection{Contrastive Learning Models}

SimCSE \cite{2104.08821} presented a simple contrastive learning framework that vastly improved the state-of-the-art sentence embeddings. By training unsupervised models using contrastive objective, the resulting quality of sentence embeddings improved by 4.2\%. The contrastive objective can be understood as maximising the lower bound of mutual information between different views of the data \cite{1808.06670}. Ultimately, the goal of contrastive learning is to learn effective representation via pulling semantically close neighbours together and pushing apart  non neighbours,  through a score function which measures similarity between these two features. This would be elaborated on the solution section. The input sentences are encoded with a pretrained language model such as DistilBERT , which is elaborated upon in the next section.

\subsection{Knowledge Distillation Models}

In our present work, contrastive learning is applied on a distilled version of BERT, DistilBERT. Sanh et. al \cite{1910.01108} was able to reduce BERT model size by 40\% while retaining 97\% of language understanding capabilities by leveraging on knowledge distillation. DistilBERT is able to retain a significant extent of model performance despite a huge reduction of memory footprint because of the benefits conferred from knowledge distillation. In knowledge distillation, DistilBERT is known as the student, while BERT is known as the teacher. Both have the same architecture. For instance, given this sample input sentence ``the \textless mask \textgreater licked its fur and howled'', the teacher model outputs ``dog'' as the mask, but ``wolf'' and ``fox'' also have high scores. The distillation loss between the teacher and the student aims to align the probability distributions, rather than just the hard predictions, which is the output from the teacher model. The probability distribution is a useful addition as the student is able to learn to rank ``wolf'' and ``dog'' highly, which is useful info about the world \cite{gatech2}.

\subsection{Combining Contrastive Learning with Knowledge Distillation}

The core key elements of contrastive learning are augmentation techniques to create positive samples, hard negative mining techniques to create negative samples, and sufficiently large batch size \cite{2002.05709}. We hypothesise that there is nothing inherent about these properties that would prevent contrastive learning’s effective use when applied to DistilBERT instead of BERT.  We believe that it would be especially meaningful to pursue this combination of contrastive learning and knowledge distillation techniques as these were the two methods that were highlighted as the most dominant and effective surrogate tasks used in unsupervised learning and self supervised learning in the lectures of the deep learning class at Georgia Tech. \cite{gatech}.

There has been some work done today to combine the two techniques.  The Contrastive Distillation on Intermediate Representations (CoDIR) framework \cite{2009.14167} uses contrastive learning to distil the teacher’s hidden states to the student through three objectives, an original training objective, a  distillation objective, and an additional contrastive learning objective where the student is trained to distil knowledge through intermediate layers of the teacher. However, this utilisation of both teacher’s output layer and its intermediate layers for student model training is a deviation from traditional knowledge distillation methods, and involves the creation of a new framework.This differs from our intention of checking compatibility of existing methods.  This framework is also instead applied directly on RoBERTa,  and is more complex to implement.  In addition the main objective of applying contrastive learning in CODIR is to compress BERT \cite{2009.14167}, whereas for us the compression is mainly done through the use of DistilBERT, and application of contrastive learning is mainly for enhancing performance.

\subsection{Pooling Methods}

The original authors of BERT \cite{1810.04805} experimented with different pooling strategies and the top three performers were concat last 4, weighted sum of last four hidden and second-to-last hidden. However, these were on fine-tuned tasks and may not be good benchmarks for semantic quality of sentence representations. Han Xiao \cite{bert.as.service} claims that the last layer is too close to the target function and is biased towards the pre-training task targets and hence, argues that the second-to-last layer is a better sentence representation. Other studies \cite{li-etal-2020-sentence} and  \cite{https://doi.org/10.48550/arxiv.1908.10084} showed that taking average of first and last layers leads to better sentence representations. In this study, we also aim to experiment with various pooling methods to contribute more data for the research community on which pooling method works best.

\section{Approach}

As DistilBERT proved to be promising in reducing memory footprint with minimal loss in language understanding, we decided to start off with an initial evaluation benchmark for BERT and DistilBERT. We evaluated these two models on Semantic Textual Similarity (STS) tasks using Spearman Correlation score as the main evaluation metric. Surprisingly, results showed that pretrained DistilBERT had consistently outperformed pretrained BERT on STS tasks. BERT scored an average of 53.73, while DistilBERT scored an average of 58.23. Hence, we think DistilBERT is a promising candidate and can be successful in achieving our goals.

We adopted a similar approach to unsupervised SimCSE \cite{2104.08821} in applying contrastive learning to pre-trained language models. To perform contrastive learning in self-supervised fashion, we take sentences \(\{x_i\}^m_{i=1}\) and use respective \(x_i\) as \(x_i^+\). During forward pass, dropouts from DistilBERT encoder is applied, resulting in embeddings that are similar but not identical for each pair of \(x_i\) and \(x_i^+\). We can denote these two embeddings as \(h_i^{z} = f(x_i, z)\) and \(h_i^{z'} = f(x_i, z')\) where \(z\) and \(z'\) are two different random dropout masks applied on \(x_i\). The cosine similarity is calculated for positive pairs \(x_i\) and \(x_i^+\), and negative pairs \(x_i\) and \(x_j^+\). Finally, our training loss for contrastive learning is computed as:

\[ \ell_i = -log \frac{e^{sim(h_i, h_i^+)/\tau}}{\sum_{j=1}^N e^{sim(h_i, h_j^+)/\tau}} \]

where \( \tau \) is a temperature hyperparameter that can be tuned for improving model performance. Here, minimizing the loss function will result in maximizing the similarity between positive pairs (increasing the numerator), and minimising the similarity between negative pairs (decreasing the denominator). Using contrastive loss function, the 66M learned parameters of DistilBERT will be tuned. The overall contrastive learning architecture using DistilBERT is shown in Figure 1 below:

\begin{figure}[hbt!]
\centering
\includegraphics[scale=0.65]{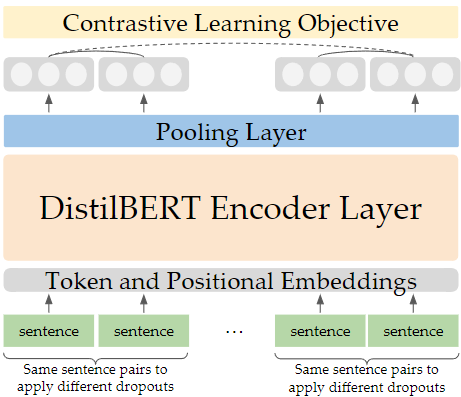}
\caption{Overall DistilFACE Architecture. Similarity of final embeddings after the pooling layer are measured. Solid lines are positive examples, while dotted lines are negative examples.}
\label{fig:short}
\end{figure}

\subsection{Training Dataset: Wiki 1M}
DistilFACE model is trained using Wiki 1M sampled dataset obtained from SimCSE. The Wiki 1M dataset was derived from the original English Wikipedia dataset through random sampling. Each row in the dataset is a sentence extracted from Wikipedia articles. This is chosen as the dataset for training because authors of SimCSE \cite{2104.08821} found that such NLI datasets are effective for learning sentence embeddings. 

\subsection{Evaluation Dataset: STS Task Datasets}

In the evaluation phase, we used STS-12, STS-13, STS-14, STS-15 and STS-B as our evaluation set for measuring model performance on semantic tasks. Each row in STS Task datasets contains a pair of sentences. Each pair is manually labelled with a score ranging from 0 to 5 to indicate how similiar or relevant the pair of sentence is to each other. We chose to use this dataset for evaluation since it has been widely used as a benchmark for STS performance.

\section{Methodology}

Pre-trained DistilBERT model with initial checkpoints from Transformers Hub is downloaded and used as the starting point for applying contrastive learning on DistilBERT.

With reference to common hyper-parameters used by previous works in BERT, DistilBERT, and SimCSE, we started with a default set of hyperparameters in this study shown in table below:

\begin{table}[hbt!]
\centering
\includegraphics[scale=0.57]{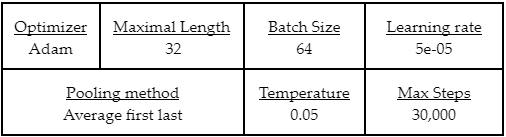}
\caption{Default Hyperparameters}
\label{fig:short}
\end{table}

A grid search is performed to find the optimal set of hyperparameters, more explanation can be found in the Results \& Analysis section. In grid search, we also perform evaluation to understand the effect of hyperparameters on STS task performance.

\subsection{Success Metrics}

Following the methodology in SimCSE, the SentEval engine by Conneau et. al. \cite{1803.05449} was mainly used to evaluate model performance on these evaluation datasets, and Spearman Correlation is used as the evaluation metric. This metric helps us to determine the quality of the sentence embeddings produced for determining semantic similarity.

To ensure our Spearman correlation results are comparable with SimCSE, the final Spearman correlation score for each dataset is calculated by first concatenating all subsets of data within the STS dataset, and then calculating Spearman correlation once, similar to the approach by SimCSE. This is slightly different from the default calculation from SentEval, which takes the average of Spearman correlation scores for each data subset within each STS dataset.

To measure success, after hyperparameter tuning is complete, the best hyperparameters will be used with max steps tuned to adjust for overfitting in the evaluation set. Spearman Correlation scores for the final DistilFACE will be compared with other models such as BERT, DistilBERT and SimCSE. 

\subsection{Further Enhancements}

To make our model even more efficient and require less GPU memory, we also adopted the use of Automatic Mixed Precision (AMP). Instead of always storing weights and biases in 32-bit float format, 16 bit is sometimes used where appropriate. This process of autocasting to lower precision where appropriate reduces wastage of GPU memory. An initial issue faced was that performance was noticeably worse. However, we realized that this was likely due to underflow, as the gradients can become too small to be stored via this format. This can be managed by applying a scaling factor to gradients. The difference made by this scaling factor increased average scores across benchmarks by 8.6\%. Overall, utilizing AMP resulted in a speed up of about 1.4x for training with a batch size of 64, to up to 3.2x for batch size of 256, and meant we needed a maximum of 8GB in GPU memory during our training process, enabling local training in line with our secondary project objectives.

We also managed to further reduce the model's size by 52\% through quantization (refer to Appendix Table 9). This involved converting a floating point model’s weights and activations to reduced precision integers post-training.  However, this reduced model performance by about 6.9\% on average. Hence, we decided to stop pursuing this trade-off further, and instead focus on our core metric of model performance.

\section{Results \& Analysis}

We compare the evaluation results of our final DistilFACE model with other models:

\begin{table}[hbt!]
\centering
\includegraphics[scale=0.4]{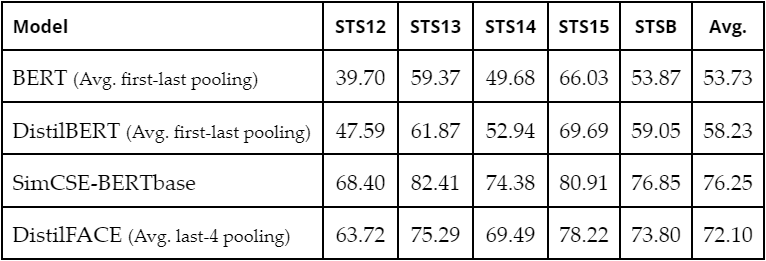}
\caption{Comparison of Model Performances. We report Spearman Correlation above as \(\rho \times 100\)}
\label{fig:short}
\end{table}

 We can see that our application of contrastive learning to DistilBERT has been clearly successful, with DistilFACE performing on average 34.2\% and 23.8\% higher in Spearman Correlation compared to BERT and DistilBERT respectively. We have also achieved our secondary goal of building a lightweight model, since DistilFACE, which is based on DistilBERT, is significantly smaller and faster than BERT \cite{1910.01108} (shown in Appendix Table 8).  In fact, DistilFACE and quantized DistilFACE are respectively 1.64 and 3.15 times smaller than the BERT implementation (shown in Appendix Table 9). 

We believe that our experiments were successful for two key reasons. The first, is that our hypothesis in section 2.3 that the properties of contrastive learning are compatible with DistilBERT is logically sound. The second, is effective hyperparameter tuning and experimentation with pooling methods enabled good performance. We detail the results and analysis of that process below.

\subsection{Hyperparameter tuning}

These are the hyperparameters that we tuned: Similarity Temperature, Batch Size, Learning Rate and Pooling Method. Each hyperparameter is tested against their selected set of values, which are adapted from \cite{2104.08821}. The optimal hyperparameters and evaluation score by training steps is shown below:

\begin{table}[hbt!]
\centering
\includegraphics[scale=0.60]{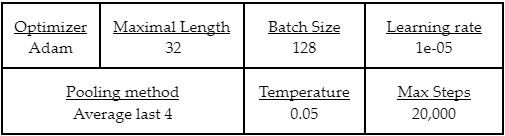}
\caption{Optimal Hyperparameters for Final DistilFACE}
\label{fig:short}
\end{table}

\begin{figure}[hbt!]
\centering
\includegraphics[scale=0.70]{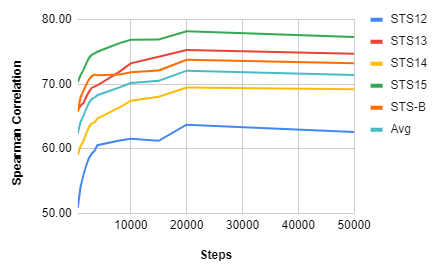}
\caption{Spearman Corr. by Steps on STS datasets}
\label{fig:short}
\end{figure}

We tuned the final model based on the max steps to avoid overfitting as well. As shown in Figure 2 above, we can see DistilFACE starts to overfit at 20,000 steps and above. Hence, the best max steps for our final model is at 20,000.

\subsection{Challenges Faced}

A problem we encountered was during experimentation with hyperparameter tuning frameworks like Ray and Optuna to enable more sophisticated methods. We did not anticipate issues as there are examples of successful applications to BERT models. However, these frameworks were found to be incompatible with our use case. Firstly, the trainer required labels from the dataset by default, which is not applicable for our unsupervised model. This caused the ``compute\_ metric'' function to be ignored by the tuning framework, which is not ideal given that we require the use of the SentEval toolkit for evaluation. Secondly, our custom model in evaluation mode does not directly output evaluation loss. Instead such metrics are calculated separately outside the model via the SentEval toolkit, which is too complex to be integrated directly.

Hence, we instead adopt the grid-search methodology adopted by the authors in \cite{2104.08821}. Each set of hyperparameter values are used in training for 30,000 steps, and then evaluated independently on these 5 datasets described above. We selected this number of steps as we found training loss to be stable by this threshold across the different hyperparameters. This reasonable amount of steps also meant that using grid search proved sufficient, as gains from sophisticated methods like parallelisation via Ray and Optuna would be more limited.

\subsection{Learning Rate}

Learning rates of 1e-5, 3e-5 and 5e-5 were selected as hyperparameter values as adapted from \cite{2104.08821}. Learning rate of 1e-5  resulted in the best average Spearman correlation, as highlighted in the yellow bar in Figure 3 and Table 4 in Appendix B. The figure also shows that smaller the learning rate, the greater the Spearman correlation.

\subsection{Similarity Temperature}

Temperatures of 0.001, 0.01, 0.05, 0.1, 1 were tested, with 0.05 achieving the best average Spearman correlation scores. (see as Figure 4 and Table 5 in Appendix B). This parameter scales the inner product between two normed feature embeddings\cite{2112.01642} when calculating the training objective. Temperature controls how sensitive the objective is to specific embedding locations \cite{2110.04403}. As such this value needs to be neither too large nor small, and in this scenario optimally at 0.05, as highlighted in the yellow bar (see Figure 4 in Appendix B).

\subsection{Batch Size}

Batch sizes of 64, 128, 256 were tested, with size 128 achieving the best average Spearman correlation scores, as highlighted in the yellow bar (see Figure 5 and Table 6 in Appendix B). However, this figure also shows that the model is not too sensitive to batch sizes if a suitable learning rate is used, consistent with the findings from \cite{2104.08821}. This further validates that contrastive learning need not require very large batch sizes \cite{2104.08821}, unlike what was previously thought\cite{2002.05709}. This also further supports our overarching hypothesis that contrastive learning is suitable to be applied to more light weight models.

\subsection{Pooling Methods}

We experimented with various pooling methods taking reduce functions: average and max, as well as the selection of hidden layers: last hidden, second-to-last, all hidden, first and last, last two, last four and concatenation of last four layers (see Figure 6 and Table 7 in Appendix B). Our findings are as follows:

\textit{Best Pooling Methods.} Concat and average of the last four layers were the best two performing pooling methods, as highlighted in the yellow bar. This is consistent with BERT \cite{1907.11692} where concatenation of the last four layers worked best for them.

\textit{Next Best Pooling Methods.} Authors \cite{li-etal-2020-sentence} \cite{https://doi.org/10.48550/arxiv.1908.10084} mentioned that taking the average of first and last leads to better results. We find this consistent with our results with average of first and last to be one of the better performing pooling methods out of all we have experimented.

\textit{Never too close.} Our results show that the last hidden is the third best performing pooling method, better than the second-to-last layer. This discredits the claim from Han Xiao \cite{bert.as.service} that the last layer is too close to pre-trained tasks and the second-to-last layer is a better sentence representation. Here, we find this to be false.

\section{Conclusion}

Overall, we believe our project to be a success using the success metrics in Section 4.1, namely Spearman Correlation on STS tasks. This can be seen from the results and rationale in Section 5. Hence we hope our DistilFACE model can help lay the groundwork for future work involving the building of lightweight models for such tasks using contrastive learning and knowledge distillation.


\small{
\bibliographystyle{bibstyle}
\bibliography{refs}
}

\begin{appendices}

\section{Project Code Repository and Results}
The Github repository for our final project is at: \url{https://github.com/kennethlimjf/contrastive-learning-in-distilled-models}.
Results can be found in Google Sheets: \href{https://docs.google.com/spreadsheets/d/1OfyPI4u9lc5NnkdpgvJFfPLUDZ8BTxlnjT0_v__VMdE/edit?usp=sharing}{STS Benchmark Results on Google Sheet}
\section{Tables and Charts}

\begin{figure}[hbt!]
\centering
\includegraphics[scale=0.3]{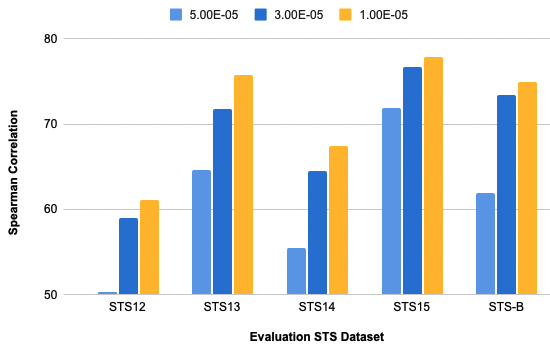}
\caption{Spearman Corr. by Learning Rate on STS datasets}
\label{fig:short}
\end{figure}

\begin{table}[hbt!]
\centering
\includegraphics[scale=0.6]{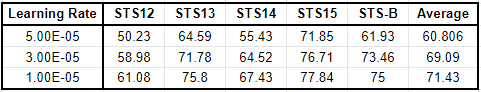}
\caption{Spearman Corr. by Learning Rate on STS datasets}
\label{fig:short}
\end{table}

\begin{figure}[hbt!]
\centering
\includegraphics[scale=0.30]{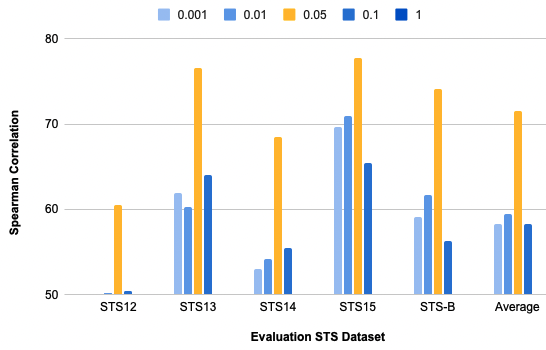}
\caption{Spearman Corr. by Similarity Temperature on STS datasets}
\label{fig:short}
\end{figure}

\begin{table}[hbt!]
\centering
\includegraphics[scale=0.55]{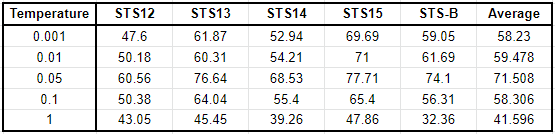}
\caption{Spearman Corr. by Similarity Temperature on STS datasets}
\label{fig:short}
\end{table}

\begin{figure}[hbt!]
\centering
\includegraphics[scale=0.35]{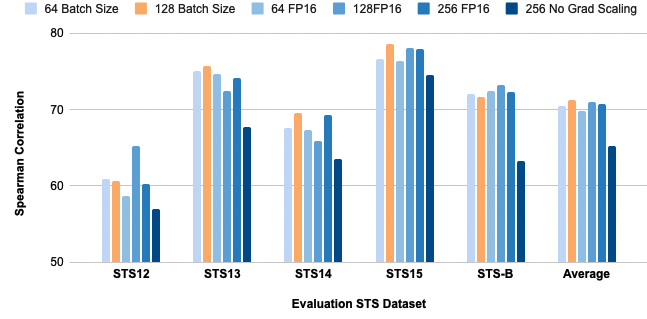}
\caption{Spearman Corr. by Batch Size on STS datasets}
\label{fig:short}
\end{figure}

\begin{table}[hbt!]
\centering
\includegraphics[scale=0.45]{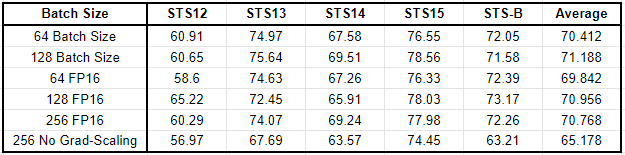}
\caption{Spearman Corr. by Batch Size on STS datasets}
\label{fig:short}
\end{table}

\begin{figure}[hbt!]
\centering
\includegraphics[scale=0.35]{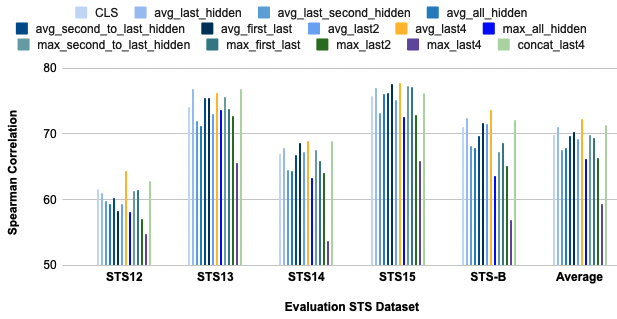}
\caption{Spearman Corr. by Pooling Methods on STS datasets}
\label{fig:short}
\end{figure}

\begin{table}[hbt!]
\centering
\includegraphics[scale=0.45]{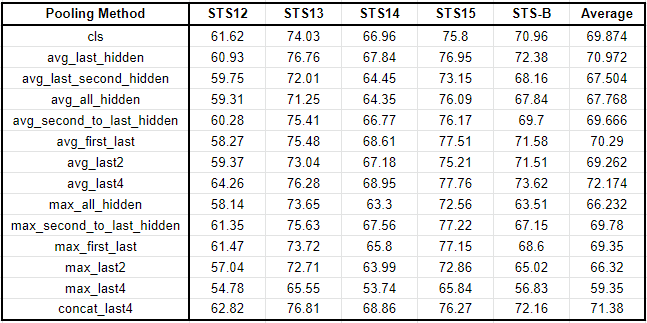}
\caption{Spearman Corr. by Pooling Methods on STS datasets}
\label{fig:short}
\end{table}

\begin{table}[hbt!]
\centering
\includegraphics[scale=0.57]{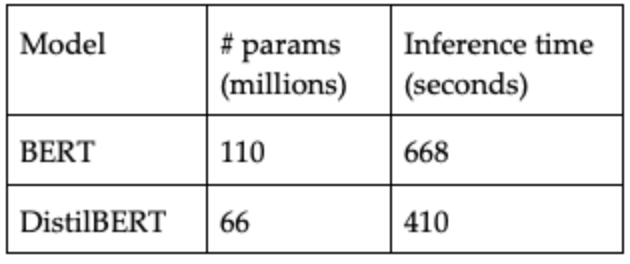}
\caption{Difference in number of parameters and inference time between BERT and DistillBERT from \cite{1910.01108}. Inference time of a full pass of STS-B on CPU with batch size of 1}
\label{fig:short}
\end{table}

\begin{table}[hbt!]
\centering
\includegraphics[scale=0.65]{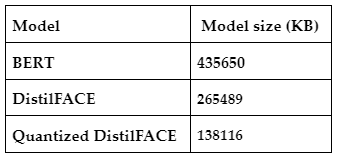}
\caption{Comparison of Model Size}
\label{fig:short}
\end{table}

\begin{table}[hbt!]
\includegraphics[scale=0.5]{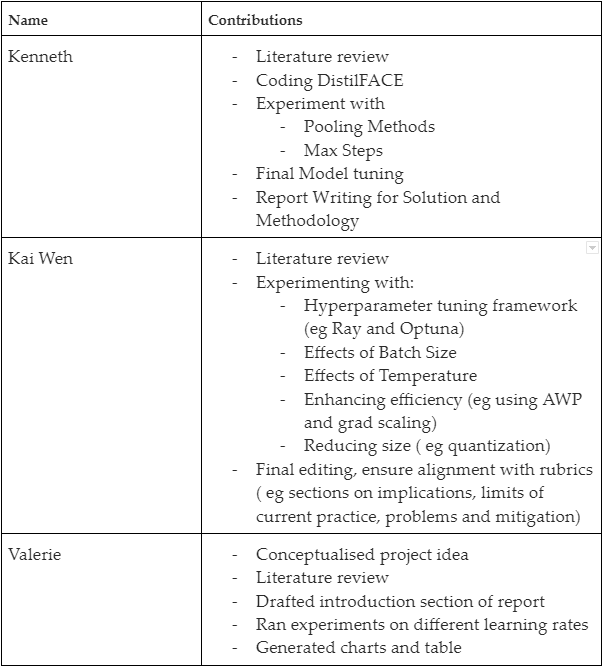}
\caption{Team member contributions}
\label{fig:short}
\end{table}

\end{appendices}

\end{document}